\pdfoutput=1

\documentclass[11pt]{article}

\usepackage[preprint]{acl}

\usepackage{times}
\usepackage{latexsym}

\usepackage[T1]{fontenc}

\usepackage[utf8]{inputenc}

\usepackage{microtype}

\usepackage{inconsolata}

\usepackage{graphicx}
\usepackage{multirow}
\usepackage{booktabs}
\usepackage{makecell}
\usepackage{graphicx} 
\usepackage{float}    
\usepackage{enumitem}
\usepackage{algorithm}
\usepackage{amsmath}
\usepackage{algpseudocode}
\usepackage[export]{adjustbox}

\usepackage{subcaption}

\newcommand{\smallsection}[1]{\noindent\textbf{#1}.}

\title{Order Matters: Rethinking Prompt Construction in In-Context Learning}

\author{
  Warren Li$^1$\thanks{Equal Contribution.}$\quad$Yiqian Wang$^{2*}\quad$Zihan Wang$^1\quad$ Jingbo Shang$^{1}$ \\
  $^1$UC San Diego $\quad$ $^2$Cushing Academy, Boston\\
  $^1$\{wyl003, ziw224, jshang\}@ucsd.edu $\quad$ $^2$yiqianwang27@cushing.org
}

\begin{document}
\maketitle
\begin{abstract}
In-context learning (ICL) enables large language models to perform new tasks by conditioning on a sequence of examples. 
Most prior work reasonably and intuitively assumes that \emph{which examples are chosen} has a far greater effect on performance than \emph{how those examples are ordered}, leading to a focus on example selection. We revisit this assumption and conduct a systematic comparison between the effect of selection and ordering.
Through controlled experiments on both classification and generation tasks, using multiple open-source model families (0.5B–27B parameters) and GPT-5, we find that the variance in performance due to different example orderings is comparable to that from using entirely different example sets. 
Furthermore, we show that strong orderings can be identified using only a development set, achieving performance close to an oracle that selects the best ordering based on test labels. 
Our findings highlight the equal and intertwined importance of example selection and ordering in prompt design, calling for a reexamination of the assumptions held in ICL.

\end{abstract}

\section{Introduction}
\label{sec:introduction}

In-context learning (ICL) has emerged as a defining capability of large language models (LLMs), enabling them to perform tasks by conditioning on a few examples provided directly in the input context—without gradient updates or task-specific fine-tuning~\cite{brown2020language}. 
This paradigm has proven remarkably effective across diverse natural language processing tasks, including text classification, question answering, and arithmetic reasoning.

It is known that the performance of ICL is sensitive to the few examples used. 
Prior works have shown that both the selection of examples in context~\cite{min2022rethinking,rubin2022learning} and the ordering of the examples affects model behavior and downstream accuracy~\cite{zhao2021fantastically}. 
Intuitively, one would expect that the ordering of examples plays a less significant role than the selection of the examples.
Indeed, much recent work has been focusing on finding a good set of examples~\cite{gao2021making,dai2022promptagator,scarlatos2023reticl}.

While previous studies have noted that order can affect accuracy, most works implicitly assume it plays a secondary role compared to selection. We revisit this assumption through a controlled study spanning both classification and generation tasks, measuring the relative variance each source contributes. Our analysis reveals that ordering and selection have comparable influence on ICL performance.
We show that varying the order of a fixed set of in-context examples leads to performance fluctuations \emph{comparable} to changing the examples themselves. 
We also explore strategies for identifying strong orderings using only an in-distribution development set, and find that such learned orderings can achieve accuracy close to the optimal ordering determined post hoc from test labels. 
However, we observe that these orderings are fragile and often fail to generalize across domains or tasks.

Our main contributions are as follows.
\begin{itemize}[leftmargin=*,nosep]
\item We quantify the performance impact of example ordering in ICL and show that it rivals the effect of example selection across multiple tasks and model sizes.
\item We propose a simple method to identify effective orderings based on a development set and evaluate its generalizability.
\item We demonstrate that no single ordering strategy consistently dominates, highlighting the instability and context dependence of prompt order effects.
\item We find that orderings discovered on one dataset rarely transfer well to another, indicating that order sensitivity is highly dataset-dependent.
\end{itemize}

These findings call for a rethinking of prompt construction in ICL, suggesting that ordering deserves as much attention as selection in efforts to improve model performance.
\section{Related Work}
\label{sec:related-works}

\smallsection{Prompt Order Sensitivity}
Several studies have documented that large language models are highly sensitive to the order of in-context examples~\cite{zhao2021fantastically, lu2022order}.
For example, \citet{zhao2021fantastically} showed that GPT-3's few-shot accuracy can vary by tens of percentage points when only the \emph{order} of examples is changed, attributing this to the model's over-reliance on early-context examples.
\citet{lu2022order} further demonstrated that optimal orderings can yield near-SOTA performance, while poor ones reduce accuracy to near-random levels across tasks like SST-2 and AGNews.

\smallsection{Example Selection vs. Ordering Effects}
While prior work emphasizes example selection as a key factor in ICL performance~\cite{min2022rethinking, rubin2022learning}, recent studies show that ordering can have a comparably large impact~\cite{zhang2022impact, guo2024demo}.
In many cases, strong examples yield weak results if poorly ordered. 
We design controlled experiments to disentangle and quantify the relative influence of selection versus ordering.

\smallsection{Strategies for Mitigating Order Sensitivity}
Heuristics like sampling permutations on a development set~\cite{zhao2021fantastically, zhang2022impact}, adaptive reordering based on the test query~\cite{guo2024demo}, and reinforcement learning-based search~\cite{bhope2023structured} have all been proposed to address order sensitivity.
Our findings highlight both the significance of ordering and the difficulty of identifying universally good permutations, motivating further work on adaptive prompt design.

\begin{table}[t]
  \centering
  \small
  \caption{Dataset statistics: number of classes \(\lvert\mathcal{C}\rvert\) and shots per prompt \(k\). All datasets use a held‐out development set \(D_{\mathrm{dev}}\) of size \(N_{\mathrm{dev}}=1000\) (see \ref{subsec:optimal-order-setup}) and a test set \(D_{\mathrm{test}}\) of size \(N_{\mathrm{test}}=500\).
  }
  \vspace{-3mm}
  \label{tab:dataset-stats}
  \begin{tabular}{cllc}
    \toprule
    \textbf{Task} & \textbf{Dataset}      & \(\lvert\mathcal{C}\rvert\) & \(k\) \\
    \midrule
    \multirow{5}{*}{\rotatebox[origin=c]{90}{Classification}} 
    & AG News               & 4   & 8  \\
    & NYT-Topics            & 9   & 18 \\
    & NYT-Locations         & 10  & 20 \\
    & DBPedia               & 14  & 28 \\
    & MMLU                  & 4   & 8  \\
    \midrule
    \multirow{3}{*}{\rotatebox[origin=c]{90}{Gen.}} 
    & GSM8K                 & —   & 8  \\
    & MMLU-Pro              & —   & 8  \\
    & MATH                  & —   & 8  \\
    \bottomrule
  \end{tabular}
  \vspace{-3mm}
\end{table}

\section{Example Selection vs. Ordering}

This section quantifies and compares two key sources of variation in few-shot in-context learning:
\begin{itemize}[nosep,leftmargin=*]
\item \textbf{Order Sensitivity}: Accuracy variation from permuting a fixed set of examples.
\item \textbf{Selection Sensitivity}: Accuracy variation from changing the example set under a fixed order.
\end{itemize}

\smallsection{Fixed Order}
Normally, a selection of examples comes with a random ordering, rendering it difficult to compare order sensitivity with selection sensitivity if the latter is compounded with the former. Therefore, we define a \textbf{default ordering} for any set of examples,
and consider a fixed ordering as a fixed permutation applied on the default ordered examples.
Specifically, to isolate the effect of ordering from example selection, we define a consistent \emph{default ordering} for each demonstration set:
\begin{itemize}[nosep,leftmargin=*]
    \item \textbf{Classification tasks:} Group examples by label, sort labels alphabetically, and sort examples within each label alphabetically.
    \item \textbf{Generation tasks:} Sort all examples alphabetically.
\end{itemize}
This ordering serves as a reference point for measuring the effects of both reordering (permutations) and replacing (selections) of examples.

\begin{figure}[t]
    \centering
    \includegraphics[width=0.9\linewidth]{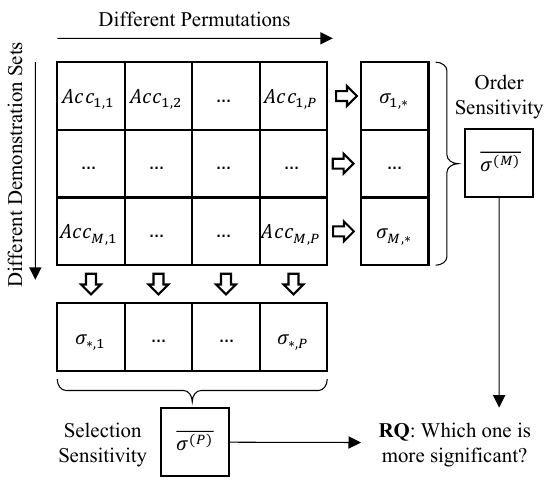}
    \vspace{-3mm}
    \caption{Measuring selection vs.\ ordering sensitivity via average grouped standard deviation.}
    \label{fig:avg_group_std}
\end{figure}

\subsection{Experiment Setup}

We evaluate both classification and generation tasks using few-shot in-context learning. We vary one variable at a time: selection or ordering, while keeping the other fixed. This design isolates their marginal effects without assuming independence between them.
For each task, we construct $M$ distinct demonstration sets \(S_1,\dots,S_M\), each containing $k$ examples. 
For every set, we evaluate $P$ random permutation $\pi_1,\ldots,\pi_P$.
All results are reported on a held-out evaluation set $D_{test}$ of 500 examples.
In this experiment, we set $M=10$ and $P=10$ for the computational consideration.

\begin{itemize}[nosep,leftmargin=*]
  \item \textbf{Classification:} 
    We use AG News~\cite{zhang2015character}, NYT-Topics~\cite{Sandhaus2008Nyt}, NYT-Locations~\cite{Sandhaus2008Nyt}, DBPedia~\cite{auer2007dbpedia}, and MMLU~\cite{hendrycks2020measuring}.
    Each dataset is balanced across classes by oversampling underrepresented labels on both $D_{dev}$ and $D_{test}$. 
    We denote the label set by \(\mathcal{C}\) and set $k = 2 |\mathcal{C}|$ for each dataset.
  \item \textbf{Generation:} 
    We evaluate on GSM8K~\cite{cobbe2021gsm8k}, MMLU-Pro~\cite{wang2024mmlupro}, and MATH~\cite{hendrycksmath2021}.
    Prompts include $k$ demonstrations followed by a free-form test query.
    Model predictions are scored using exact match or numeric tolerance.
\end{itemize}
Dataset statistics, including number of classes and shots per prompt, are summarized in Table~\ref{tab:dataset-stats}.

\begin{figure}[t]
  \centering
  \includegraphics[width=1.0\linewidth]{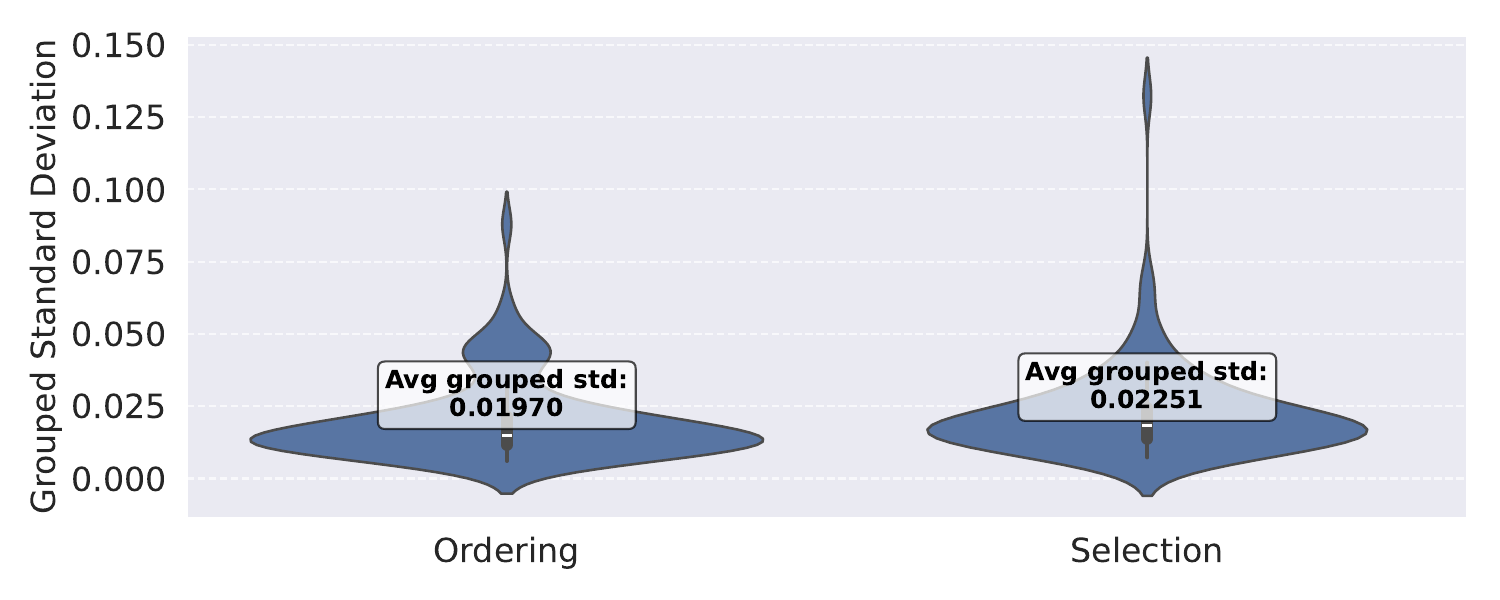}
  \vspace{-3mm}
  \caption{
  Violin plots of the general distributions of ordering sensitivity and selection sensitivity across all model–dataset combinations. 
  }
  \label{fig:sensitivity_violin_plot}
\end{figure}

\subsection{Sensitivity Metrics}

Figure~\ref{fig:avg_group_std} illustrates our two sensitivity metrics, both computed as \emph{average grouped standard deviations} over $M$ demonstration sets and $P$ permutations.

Let $a_{i,j} = Acc(S_i,\pi_j\mid D_{test})$ denote the accuracy on evaluation set $D$ when using the $j$-th permutation of the $i$-th demonstration set.

\smallsection{Order Sensitivity}
We compute the standard deviation across permutations for each demonstration set (row-wise), then average over all $M$ sets:
\begin{equation}
\overline{\sigma^{(M)}} = \frac{1}{M}\sum_{i=1}^M \mathrm{std}{a_{i,1},\dots,a_{i,P}}.
\end{equation}

\smallsection{Selection Sensitivity}
We compute the standard deviation across demonstration sets for each permutation (column-wise), then average over all $P$ permutations:
\begin{equation}
\overline{\sigma^{(P)}} = \frac{1}{P}\sum_{j=1}^P \mathrm{std}{a_{1,j},\dots,a_{M,j}}.
\end{equation}

\smallsection{Relative Importance}
We define the ratio
\begin{equation}
    r = \overline{\sigma^{(P)}} / \overline{\sigma^{(M)}}
\end{equation}
to compare the relative impact of selection vs. ordering. A ratio near 1 implies equal importance, $r > 1$ indicates selection dominates, and $r < 1$ implies ordering has greater influence.

\subsection{Results}

\smallsection{Overall}
Figure~\ref{fig:sensitivity_violin_plot} shows the violin plots for the general distributions of ordering sensitivity and selection sensitivity. 
The two distributions are very similar to each other, indicating their comparable impact.
Remarkably, the average selection sensitivity is $0.02251$, which is only $\sim$14\%~ more than the average ordering sensitivity $0.01970$.
These results indicate that the ordering of the demonstrations in ICL can be as important as the selection of the examples.

\begin{table}[t]
\small
\centering
\caption{Order vs. Selection Sensitivity per Model.}
\label{tab:sensitivity-agg-by-model}
\vspace{-3mm}
\begin{tabular}{llccc}
\toprule
\textbf{Model} & \textbf{Size}
 & \textbf{Order}
 & \textbf{Selection}
 & $r$ \\
\midrule

\multirow{4}{*}{Qwen2.5} & 0.5B
 & 0.0223
 & 0.0245
 & 1.10 \\

 & 1.5B
 & 0.0134
 & 0.0146
 & 1.10 \\

 & 3B
 & 0.0129
 & 0.0166
 & 1.29 \\

 & 7B
 & 0.0119
 & 0.0155
 & 1.30 \\
\midrule

\multirow{2}{*}{Gemma-2} & 2B
 & 0.0204
 & 0.0209
 & 1.02 \\

 & 9B
 & 0.0115
 & 0.0145
 & 1.27 \\
\midrule

\multirow{2}{*}{Gemma} & 2B
 & 0.0217
 & 0.0279
 & 1.29 \\

 & 7B
 & 0.0161
 & 0.0170
 & 1.06 \\
\midrule

\multirow{3}{*}{Llama 3} & 1B
 & 0.0184
 & 0.0200
 & 1.09 \\

 & 3B
 & 0.0152
 & 0.0172
 & 1.13 \\

 & 8B
 & 0.0277
 & 0.0393
 & 1.42 \\
\midrule

\multirow{2}{*}{DeepSeek Distill} & 1.5B
 & 0.0362
 & 0.0364
 & 1.01 \\

 & 7B
 & 0.0287
 & 0.0281
 & 0.98 \\
\midrule

\multirow{1}{*}{Gemma-3} & 27B
 & 0.0157
 & 0.0262
 & 1.67 \\
\midrule

\multirow{1}{*}{GPT-5} & Nano
 & 0.0234
 & 0.0198
 & 0.85 \\

\bottomrule
\end{tabular}%
\end{table}

\begin{figure*}[t]
    \centering
    \begin{subfigure}[b]{0.24\textwidth}
        \centering
        \includegraphics[width=\textwidth]{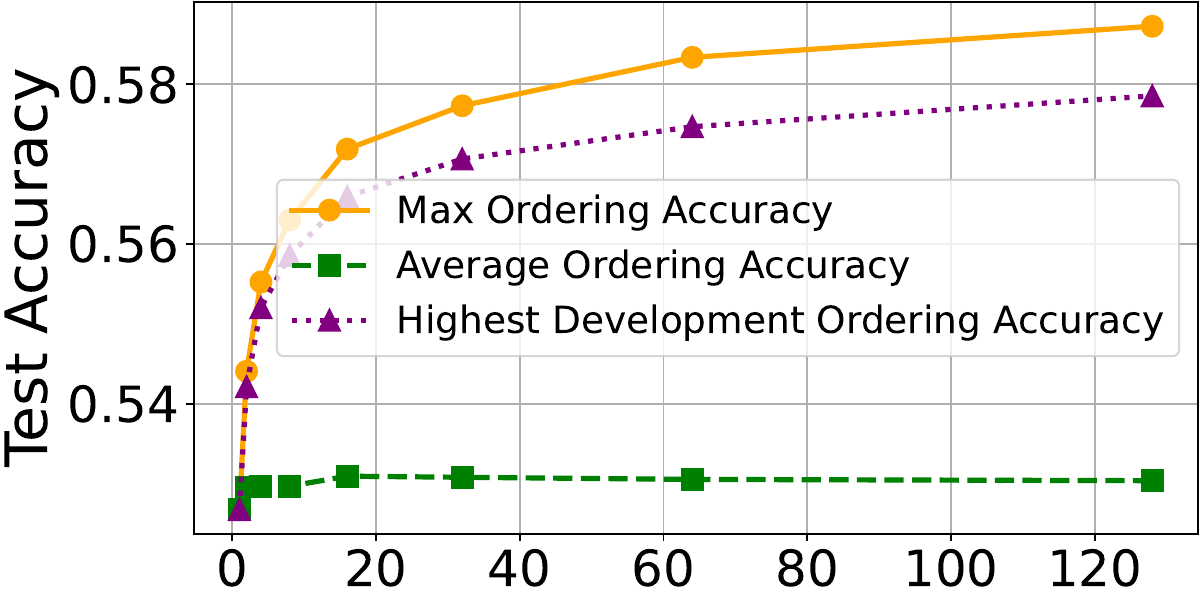}
        \caption{$P$ in Classification}
        \label{fig:aggregate_classification_subsample}
    \end{subfigure}
    \hfill
    \begin{subfigure}[b]{0.24\textwidth}
        \centering
        \includegraphics[width=\textwidth]{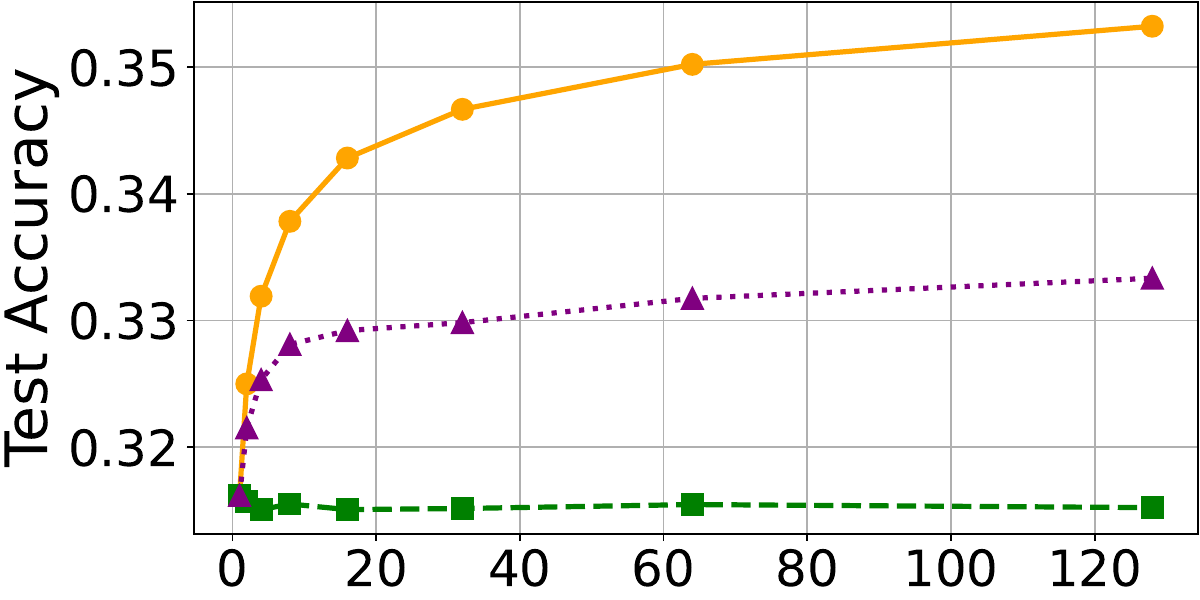}
        \caption{$P$ in Generation}
        \label{fig:aggregate_generation_subsample}
    \end{subfigure}
    \vspace{0.5em}
    \begin{subfigure}[b]{0.24\textwidth}
        \centering
        \includegraphics[width=\textwidth]{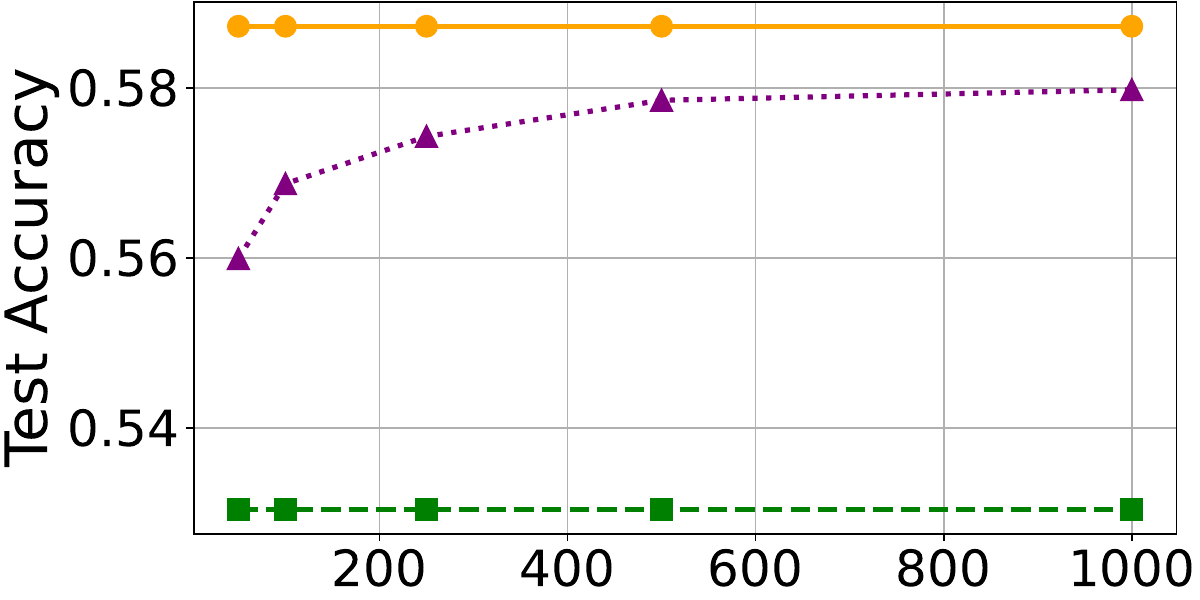}
        \caption{$|D_{dev}|$ in Classification}
        \label{fig:aggregate_classification_train_size}
    \end{subfigure}
    \hfill
    \begin{subfigure}[b]{0.24\textwidth}
        \centering
        \includegraphics[width=\textwidth]{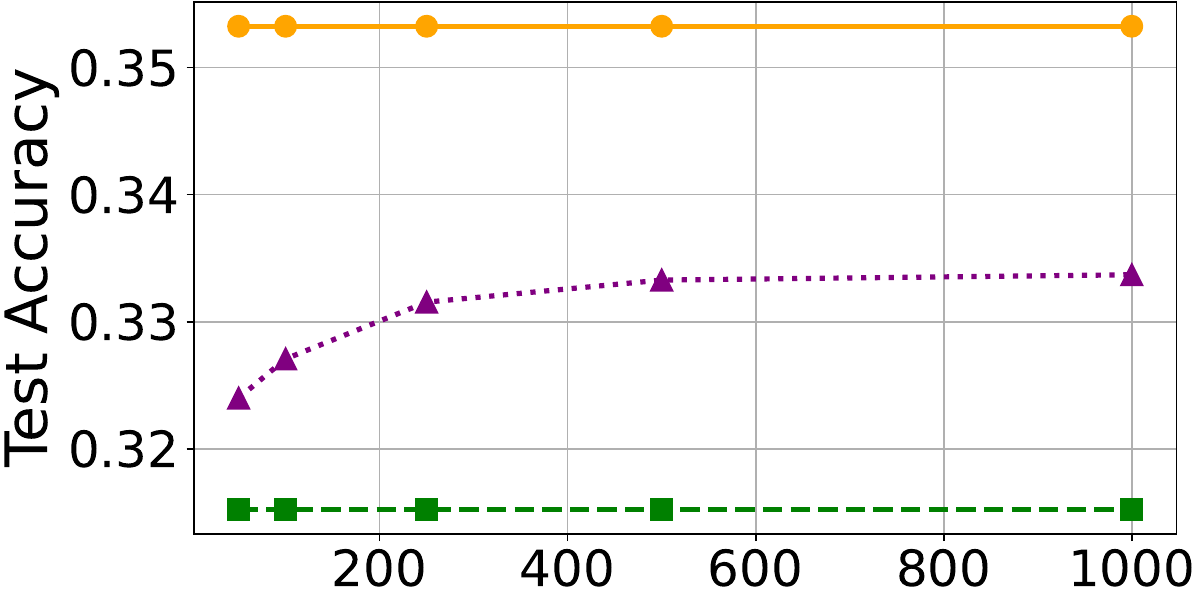}
        \caption{$|D_{dev}|$ in Generation}
        \label{fig:aggregate_generation_train_size}
    \end{subfigure}
    \caption{\textbf{Max} test accuracy (orange), \textbf{highest‐dev} test accuracy (purple), and \textbf{average} test accuracy (green) w.r.t. different parameter values. The scores are aggregated over classification and generation tasks.}
    \label{fig:finding_order}
\end{figure*}

\smallsection{Sensitivity by Model (Size)} 
Table~\ref{tab:sensitivity-agg-by-model} presents the aggregated results by model.
Across all architectures (0.5B–27B parameters, GPT-5-nano), mean accuracies under ordering and selection perturbations differ by less than one percentage point, indicating comparable central tendencies. 
The grouped standard deviations reveal a clear trend for both ordering and selection sensitivity: smaller models exhibit marginally higher variability under random permutations, while larger models remain more stable. 
This suggests an inverse relationship between model size and susceptibility to prompt order and example selection. 
However, the ratio $r$ shows no consistent monotonic trend as size increases. For some families of models, $r$ appears to be decreasing, implying ordering becomes relatively more important, whereas for others it increases. In fact, the enterprise GPT-5-nano model seems to be the most sensitive to ordering compared to all the other smaller models.

\begin{table}[t]
  \centering
  \small
  \caption{Order vs.\ Selection Sensitivity per Dataset (updated to include Gemma-3-27B and GPT-5-Nano).}
  \label{tab:sensitivity-agg-by-dataset}
  \vspace{-3mm}
  \begin{tabular}{clccc}
    \toprule
                       & \textbf{Dataset} & \textbf{Order} & \textbf{Selection} & $r$ \\
    \midrule
    \multirow{5}{*}{\rotatebox[origin=c]{90}{Classification}}
      & AG News         & 0.0273 & 0.0275 & 1.01 \\
      & DBPedia         & 0.0268 & 0.0308 & 1.15 \\
      & MMLU            & 0.0172 & 0.0206 & 1.20 \\
      & NYT Locations   & 0.0189 & 0.0216 & 1.14 \\
      & NYT Topics      & 0.0229 & 0.0223 & 0.97 \\
    \midrule
    \multirow{2}{*}{\rotatebox[origin=c]{90}{Gen.}}
      & GSM8K           & 0.0139 & 0.0220 & 1.58 \\
      & MATH            & 0.0168 & 0.0223 & 1.33 \\
    \bottomrule
  \end{tabular}
  \vspace{-3mm}
\end{table}

\smallsection{Sensitivity by Task and Dataset}  
Table~\ref{tab:sensitivity-agg-by-dataset} presents the aggregated results by dataset.
For closed‐set classification benchmarks, \(r\) lies between 0.93 and 1.20 (mean=1.09), indicating that selection sensitivity exceeds ordering sensitivity by only 9\%.

By contrast, for open-ended generation tasks, \(r\) is 1.46, exhibiting 46\% more sensitivity to selection than ordering.
Overall, generation tasks are more sensitive to selection changes relative to ordering changes when compared to classification tasks.

For the complete set of breakdowns by model and dataset, see Appendix~\ref{subsec:appendix-sensitivity}.
\section{Finding Near‐Optimal Ordering}

In prior sections, we showed that the order of in-context examples can significantly affect performance -- sometimes as much as the examples themselves. 
This naturally raises the question: can we identify a \textit{good} ordering using only a small development set, without relying on oracle access to the test set? 
In this section, we investigate whether it is feasible to search over candidate permutations and select one that generalizes well to unseen examples.

\subsection{Experiment Setup}
\label{subsec:optimal-order-setup}

We evaluate each ordering strategy using a held-out \emph{test set} \( D_{\text{test}} \) of \( N_{\text{test}} = 500 \) examples, and a \emph{development set} \( D_{\text{dev}} \) of size \( N_{\text{dev}} = 1000 \) by default.  
For each of \( P = 128 \) randomly sampled permutations of a fixed demonstration set, we compute its average accuracy on the development set across \( M = 10 \) distinct example sets.

The permutation with the highest average dev accuracy is selected and then evaluated on the test set—this result is denoted as \textbf{highest-dev}. We compare it against:
(1) \textbf{Average} test accuracy across all \( P \) permutations;
and (2) \textbf{Max} test accuracy among all \( P \) permutations (an oracle upper bound, included for reference).

We vary both \( P \) and \( |D_{\text{dev}}| \) to assess how the number of permutations and the size of the development set influence the ability to find a near-optimal ordering.

\subsection{Results}

Figures~\ref{fig:aggregate_classification_subsample} and~\ref{fig:aggregate_generation_subsample} show results for classification and generation tasks as \( P \) increases. In both cases, the \textbf{highest-dev} curve (purple) rises rapidly to match the \textbf{max} curve (orange), while the \textbf{average} (green) stays well below.

\smallsection{Classification vs. Generation}
For classification tasks, evaluating just 16–32 permutations recovers over 98\% of the oracle accuracy, reaching 99\% with 128 permutations.  
Generation tasks are more challenging, but still achieve over 95\% of the oracle within 64–100 permutations. Performance also stabilizes after 250$\sim$ dev examples in both settings.

\smallsection{Effect of Development Set Size}
Figures~\ref{fig:aggregate_classification_train_size} and~\ref{fig:aggregate_generation_train_size} vary \( |D_{\text{dev}}| \) from 50 to 1000. In both task families, performance plateaus after around 250 examples, showing that even small dev sets suffice for robust ordering selection.

\smallsection{Ordering Transfer across Datasets}
To test whether “good” orderings generalize across tasks, we evaluate transfer performance between GSM8K and MATH (8-shot generation). For each dataset, we reuse the ordering that performed best on another dataset’s development set.
As shown in Table 7, transferring the best ordering from one dataset to another results in test accuracy similar to the average random ordering for the target dataset. This indicates that ordering benefits are dataset specific and do not generalize well.
These results reinforce that while strong orderings can be learned from a development set, they are fragile and tied to the domain distribution.

\smallsection{Summary}
These results yield two key takeaways:  
(1) Our method generalizes across task types, and  
(2) Near-optimal orderings can be discovered using a modest development set—without access to test labels.
(3) Orderings are hard to transfer across different datasets

For the detailed breakdown by model and dataset, please see Appendix~\ref{subsec:appendix-optimal-order}.

\section{Conclusions and Future Work}
\label{sec:Conclusions}

In this work, we showed that the variability induced by reordering the same set of examples is comparable to the variability resulting from selecting entirely different example sets in both classification and generation tasks. Further, we demonstrated that evaluating just 64–128 candidate orderings on a small development set suffices to recover prompt permutations that achieve most of the oracle test accuracy.

Future work should test whether these findings hold for larger, state-of-the-art models like GPT-5, explore other languages and shot regimes, and extend to new task formats such as code generation and retrieval-augmented generation.


\section{Limitations}

We restrict our experiments to open‐source models in the 0.5–27B parameter range and GPT-5-nano. We do not include the leading commercial LLMs such as GPT-5 (full size) and Claude models. Expanding to those larger models would require additional GPU compute and API access, which were beyond our current budget and time constraints.

Our evaluation covers only English language classification and generation benchmarks using accuracy as the sole metric. We do not assess other languages and task types (e.g., code generation, retrieval-augmented generation, dialogue).

\section{Ethical Considerations}
We used OpenAI’s GPT-o3 and o4 models to assist with writing code and editing the paper. All technical design, experiments, and final manuscript revisions were performed and verified by the authors.

\bibliography{custom}

@inproceedings{rubin2022learning,
  title={Learning To Retrieve Prompts for In-Context Learning},
  author={Rubin, Ohad and Herzig, Jonathan and Berant, Jonathan},
  booktitle={Proceedings of the 2022 Conference of the North American Chapter of the Association for Computational Linguistics: Human Language Technologies},
  pages={2655--2671},
  year={2022}
}

@inproceedings{bhope2023structured,
  author    = {Bhope, Neha and Iyer, Anirudh and Srikumar, Vivek},
  title     = {Structured Prompt Ordering for Neural API Generation},
  booktitle = {Proceedings of the 61st Annual Meeting of the Association for Computational Linguistics (ACL)},
  year      = {2023},
  pages     = {1234--1245},
  address   = {Toronto, Canada},
  note      = {Forthcoming},
}

@inproceedings{brown2020language,
  author    = {Brown, Tom and Mann, Benjamin and Ryder, Nick and Subbiah, Melanie and Kaplan, Jared and Dhariwal, Prafulla and Neelakantan, Arvind and Shyam, Pranav and Sastry, Girish and Askell, Amanda and Agarwal, Sandhini and Herbert-Voss, Ariel and Krueger, Gretchen and Henighan, Tom and Child, Rewon and Ramesh, Aditya and Ziegler, Daniel and Wu, Jeffrey and Winter, Clemens and Hesse, Christopher and Chen, Mark and Sigler, Eric and Litwin, Mateusz and Gray, Scott and Chess, Benjamin and Clark, Jack and Berner, Christopher and McCandlish, Sam and Radford, Alec and Sutskever, Ilya and Amodei, Dario},
  title     = {Language Models are Few-Shot Learners},
  booktitle = {Advances in Neural Information Processing Systems},
  year      = {2020},
  pages     = {1877--1901},
  note      = {NeurIPS 2020},
}

@inproceedings{guo2024demo,
  author    = {Guo, Chao and Chaudhary, Aditi and Xu, Wei},
  title     = {{DEmO}: Dynamic Example Ordering for In-Context Learning},
  booktitle = {Proceedings of the 62nd Annual Meeting of the Association for Computational Linguistics (ACL)},
  year      = {2024},
  pages     = {156--169},
  address   = {Buenos Aires, Argentina},
  note      = {Forthcoming},
}

@inproceedings{lu2022order,
  author    = {Lu, Yifan and Grokhovsky, Isabella and Liu, Alexander},
  title     = {Order Matters: Re-evaluating Few-Shot Prompting for Text Classification Tasks},
  booktitle = {Proceedings of the 60th Annual Meeting of the Association for Computational Linguistics (ACL)},
  year      = {2022},
  pages     = {2345--2356},
  address   = {Dublin, Ireland},
}

@inproceedings{min2022rethinking,
  author    = {Min, Sewon and Lewis, Mike and Hajishirzi, Hannaneh and Zettlemoyer, Luke},
  title     = {Rethinking the Role of Demonstrations: What Makes In-Context Learning Work?},
  booktitle = {Proceedings of the 60th Annual Meeting of the Association for Computational Linguistics (ACL)},
  year      = {2022},
  pages     = {375--384},
  address   = {Dublin, Ireland},
}

@inproceedings{zhang2022impact,
  author    = {Zhang, Ruiqi and Gupta, Vivek and Roth, Dan},
  title     = {The Impact of Example Selection and Order on Few-Shot Language Model Performance},
  booktitle = {Findings of the Association for Computational Linguistics (ACL Findings)},
  year      = {2022},
  pages     = {945--957},
  address   = {Online and Dublin, Ireland},
}

@inproceedings{zhao2021fantastically,
  author    = {Zhao, Xinran and Wallace, Eric and Feng, Shi and Schoelkopf, Hailey and Singh, Sameer and Gardner, Matt},
  title     = {Fantastically Ordered Prompts and Where to Find Them: Overcoming Few-Shot Prompt Order Sensitivity},
  booktitle = {Proceedings of the 59th Annual Meeting of the Association for Computational Linguistics (ACL)},
  year      = {2021},
  pages     = {6193--6199},
  address   = {Online and Bangkok, Thailand},
}

@inproceedings{gao2021making,
  title     = {Making Pre-trained Language Models Better Few-Shot Learners},
  author    = {Gao, Tianyu and Fisch, Adam and Chen, Danqi},
  booktitle = {Proceedings of the 59th Annual Meeting of the Association for Computational Linguistics (ACL)},
  pages     = {5712--5726},
  year      = {2021},
  url       = {https://aclanthology.org/2021.acl-long.295}
}

@article{dai2022promptagator,
  title   = {Promptagator: Few-Shot Dense Retrieval From 8 Examples},
  author  = {Dai, Zhuyun and Zhao, Vincent Y. and Ma, Ji and Luan, Yi and Ni, Jianmo and Lu, Jing and Bakalov, Anton and Guu, Kelvin and Hall, Keith B. and Chang, Ming-Wei},
  journal = {arXiv preprint arXiv:2209.11755},
  year    = {2022},
  url     = {https://arxiv.org/abs/2209.11755}
}

@article{scarlatos2023reticl,
  title   = {RetICL: Sequential Retrieval of In-Context Examples with Reinforcement Learning},
  author  = {Scarlatos, Alexander and Lan, Andrew},
  journal = {arXiv preprint arXiv:2305.14502},
  year    = {2023},
  url     = {https://arxiv.org/abs/2305.14502}
}

@inproceedings{zhang2015character,
  title={Character-level convolutional networks for text classification},
  author={Zhang, Xiangyang and Zhao, Junbo and LeCun, Yann},
  booktitle={Advances in Neural Information Processing Systems},
  volume={28},
  year={2015}
}

@article{Sandhaus2008Nyt,
  title={The New York Times Annotated Corpus},
  author={Sandhaus, Evan},
  journal={Linguistic Data Consortium, Philadelphia},
  year={2008},
  note={LDC2008T19}
}

@inproceedings{auer2007dbpedia,
  title={DBpedia: A nucleus for a web of open data},
  author={Auer, S{\"o}ren and Bizer, Christian and Kobilarov, Georgi and Lehmann, Jens and Cyganiak, Richard and Ives, Zachary},
  booktitle={In: Proceedings of ISWC/ASWC},
  series={Lecture Notes in Computer Science},
  volume={4825},
  pages={722--735},
  year={2007},
  publisher={Springer}
}

@article{hendrycks2020measuring,
  title={Measuring massive multitask language understanding},
  author={Hendrycks, Dan and Burns, Collin and Basart, Steven and Zou, Andy and Mazeika, Mantas and Song, Dawn and Steinhardt, Jacob},
  journal={arXiv preprint arXiv:2009.03300},
  year={2020}
}

@inproceedings{wang2024mmlupro,
  title     = {MMLU-Pro: A More Robust and Challenging Multi-Task Language Understanding Benchmark},
  author    = {Wang, Yubo and Ma, Xueguang and Zhang, Ge and Ni, Yuansheng and Chandra, Abhranil and Guo, Shiguang and Ren, Weiming and Arulraj, Aaran and He, Xuan and Jiang, Ziyan and Li, Tianle and Ku, Max and Wang, Kai and Zhuang, Alex and Fan, Rongqi and Yue, Xiang and Chen, Wenhu},
  booktitle = {NeurIPS 2024 Track on Datasets and Benchmarks (Spotlight)},
  year      = {2024},
  url       = {https://arxiv.org/abs/2406.01574},
  doi       = {10.48550/arXiv.2406.01574}
}

@article{cobbe2021gsm8k,
  title={Training verifiers to solve math word problems},
  author={Cobbe, Karl and Kosaraju, Vineet and Bavarian, Mohammad and Chen, Mark and Jun, Heewoo and Kaiser, Lukasz and Plappert, Matthias and Tworek, Jerry and Hilton, Jacob and Nakano, Reiichiro and Hesse, Christopher and Schulman, John},
  journal={arXiv preprint arXiv:2110.14168},
  year={2021}
}

@article{hendrycksmath2021,
  title={Measuring mathematical problem solving with the MATH dataset},
  author={Hendrycks, Dan and Burns, Collin and Kadavath, Saurav and Arora, Akul and Basart, Steven and Tang, Eric and Song, Dawn and Steinhardt, Jacob},
  journal={NeurIPS},
  year={2021}
}

@misc{qwen2.5-tr,
  title        = {Qwen2.5 Technical Report},
  author       = {{Qwen Team}},
  howpublished = {arXiv preprint arXiv:2412.15115},
  year         = {2024}
}

@article{gemma2,
  title        = {Gemma 2: Improving Open Language Models at a Practical Size},
  author       = {{Gemma Team}},
  journal      = {arXiv preprint arXiv:2408.00118},
  year         = {2024}
}

@article{gemma1,
  title        = {Gemma: Open Models Based on Gemini Research and Technology},
  author       = {{Gemma Team}},
  howpublished = {arXiv preprint arXiv:2403.08295},
  year         = {2024}
}

@article{llama3,
  title={The llama 3 herd of models},
  author={Dubey, Abhimanyu and Jauhri, Abhinav and Pandey, Abhinav and Kadian, Abhishek and Al-Dahle, Ahmad and Letman, Aiesha and Mathur, Akhil and Schelten, Alan and Yang, Amy and Fan, Angela and others},
  journal={arXiv e-prints},
  pages={arXiv--2407},
  year={2024}
}

@article{deepseek-r1,
  title        = {DeepSeek-R1: Incentivizing Reasoning Capability in LLMs},
  author       = {Deepseek-AI Team},
  howpublished = {arXiv preprint arXiv:2501.12948},
  year         = {2025}
}

\clearpage
\newpage
\section{Appendix}
\label{sec:appendix}

Below, we include three tables summarizing our core experimental results, and the pseudocode for the development‐selected prompt ordering procedure.

\subsection{Detailed Selection Sensitivity by Model-Dataset}
\label{subsec:appendix-sensitivity}
Table~\ref{tab:main_table} summarizes the ordering and selection sensitivity for all model and dataset pairs using $M=10$ demonstration sets and $P=10$ permutations.

\begin{table*}[ht!]
\centering
\caption{Updated Average Grouped Std Dev by Model and Dataset (Ordering Sensitivity vs.\ Selection Sensitivity). We evaluate Qwen2.5~\cite{qwen2.5-tr}, Gemma 2~\cite{gemma2}, Gemma~\cite{gemma1}, LLaMA 3~\cite{llama3} and DeepSeek‐R1 Distill~\cite{deepseek-r1}.}
\label{tab:main_table}
\resizebox{\textwidth}{!}{%
\begin{tabular}{ll|cc|cc|cc|cc|cc|cc|cc}
\toprule
\multirow{2}{*}{\textbf{Model}} & \multirow{2}{*}{\textbf{Size}}
  & \multicolumn{2}{c|}{\textbf{ag\_news}} 
  & \multicolumn{2}{c|}{\textbf{dbpedia}} 
  & \multicolumn{2}{c|}{\textbf{gsm8k}} 
  & \multicolumn{2}{c|}{\textbf{math}}
  & \multicolumn{2}{c|}{\textbf{mmlu}} 
  & \multicolumn{2}{c|}{\textbf{nyt-locations}} 
  & \multicolumn{2}{c}{\textbf{nyt-topics}} \\
& & \textbf{Ordering} & \textbf{Selection}
  & \textbf{Ordering} & \textbf{Selection}
  & \textbf{Ordering} & \textbf{Selection}
  & \textbf{Ordering} & \textbf{Selection}
  & \textbf{Ordering} & \textbf{Selection}
  & \textbf{Ordering} & \textbf{Selection}
  & \textbf{Ordering} & \textbf{Selection} \\
\midrule

\multirow{4}{*}{Qwen2.5} & 0.5B 
  & 0.04581 & 0.03704 
  & 0.00835 & 0.01783 
  & 0.01158 & 0.01752 
  & 0.01258 & 0.01502 
  & 0.00996 & 0.01384 
  & 0.02287 & 0.02183 
  & 0.04490 & 0.04842 \\

& 1.5B 
  & 0.00911 & 0.00729 
  & 0.02336 & 0.02324 
  & 0.01154 & 0.01532 
  & 0.01821 & 0.02304 
  & 0.01051 & 0.01358 
  & 0.00874 & 0.00935 
  & 0.01213 & 0.01068 \\

& 3B 
  & 0.01203 & 0.01456 
  & 0.01208 & 0.01397 
  & 0.01626 & 0.03585 
  & 0.01162 & 0.01363 
  & 0.01307 & 0.01220 
  & 0.00604 & 0.01021 
  & 0.01896 & 0.01601 \\

& 7B 
  & 0.01188 & 0.01120 
  & 0.01185 & 0.01190 
  & 0.01185 & 0.02471 
  & 0.01591 & 0.02234 
  & 0.00677 & 0.01045 
  & 0.00885 & 0.01138 
  & 0.01597 & 0.01634 \\

\midrule

\multirow{2}{*}{Gemma-2} & 2B 
  & 0.04655 & 0.03853 
  & 0.02022 & 0.02114 
  & 0.01306 & 0.01836 
  & 0.01314 & 0.01920 
  & 0.00953 & 0.00909 
  & 0.01942 & 0.02089 
  & 0.02077 & 0.01903 \\

& 9B 
  & 0.00856 & 0.01091 
  & 0.00730 & 0.01201 
  & 0.01501 & 0.02188 
  & 0.01351 & 0.02068 
  & 0.00759 & 0.00935 
  & 0.00707 & 0.00862 
  & 0.02121 & 0.01839 \\

\midrule

\multirow{2}{*}{Gemma} & 2B 
  & 0.05299 & 0.06901 
  & 0.01917 & 0.03062 
  & 0.01252 & 0.02081 
  & 0.01276 & 0.01994 
  & 0.00953 & 0.01081 
  & 0.02165 & 0.02174 
  & 0.02318 & 0.02243 \\

& 7B 
  & 0.01033 & 0.01194 
  & 0.01501 & 0.01369 
  & 0.01595 & 0.01974 
  & 0.01397 & 0.01729 
  & 0.01483 & 0.01620 
  & 0.02131 & 0.02031 
  & 0.02106 & 0.01971 \\

\midrule

\multirow{3}{*}{Llama 3} & 1B 
  & 0.03845 & 0.03131 
  & 0.01758 & 0.02401 
  & 0.01524 & 0.02790 
  & 0.01720 & 0.01966 
  & 0.00801 & 0.00880 
  & 0.01269 & 0.01222 
  & 0.01992 & 0.01619 \\

& 3B 
  & 0.01239 & 0.01454 
  & 0.01036 & 0.01106 
  & 0.01441 & 0.01827 
  & 0.01419 & 0.01581 
  & 0.01071 & 0.01138 
  & 0.01527 & 0.01755 
  & 0.02892 & 0.03190 \\

& 8B 
  & 0.00995 & 0.01435 
  & 0.08791 & 0.13226 
  & 0.01070 & 0.01507 
  & 0.01013 & 0.01372 
  & 0.01014 & 0.01312 
  & 0.04282 & 0.06237 
  & 0.02222 & 0.02420 \\

\midrule

\multirow{2}{*}{DeepSeek Distill} & 1.5B 
  & 0.05886 & 0.05018 
  & 0.04405 & 0.03755 
  & 0.01842 & 0.03091 
  & 0.03005 & 0.03766 
  & 0.01908 & 0.02483 
  & 0.03914 & 0.04010 
  & 0.04365 & 0.03331 \\

& 7B 
  & 0.04589 & 0.03200 
  & 0.03366 & 0.02878 
  & 0.01485 & 0.02307 
  & 0.04383 & 0.04848 
  & 0.01313 & 0.01753 
  & 0.02615 & 0.02871 
  & 0.02316 & 0.01824 \\

\midrule

\multirow{1}{*}{Gemma-3} & 27B 
  & 0.00697 & 0.00771 
  & 0.00545 & 0.00828 
  & 0.02037 & 0.04427 
  & 0.00643 & 0.00807 
  & 0.05288 & 0.09270 
  & 0.00972 & 0.01356 
  & 0.00802 & 0.00849 \\

\midrule

\multirow{1}{*}{GPT-5} & Nano 
  & 0.03792 & 0.03026 
  & 0.06127 & 0.04807 
  & 0.00976 & 0.00937 
  & 0.02178 & 0.01577 
  & 0.01090 & 0.01087 
  & 0.00916 & 0.01095 
  & 0.01287 & 0.01359 \\

\bottomrule
\end{tabular}%
}
\end{table*}

\subsection{Finding Optimal Order on Classification and Generation Tasks}
\label{subsec:appendix-optimal-order}
Tables~\ref{tab:classification-tasks} and \ref{tab:generation-tasks} shows the Average accuracy, Highest Development selected ordering, and Max Ordering accuracy on the Classification and Generation tasks respectively at the maximum of $M=10$ demonstration sets and $P=128$ permutations.

\begin{table*}[ht!]
\centering
\caption{Classification Tasks: Accuracy metrics for \textbf{nyt-topics}, \textbf{nyt-locations}, and \textbf{dbpedia} (Average, Highest Dev, Max).}
\label{tab:classification-tasks}
\resizebox{\textwidth}{!}{%
\begin{tabular}{l
                ccc
                ccc
                ccc}
\toprule
& \multicolumn{3}{c}{\textbf{nyt-topics}}
& \multicolumn{3}{c}{\textbf{nyt-locations}}
& \multicolumn{3}{c}{\textbf{dbpedia}} \\
\cmidrule(lr){2-4}
\cmidrule(lr){5-7}
\cmidrule(lr){8-10}
\textbf{Model}
& \textbf{Average} & \textbf{Highest Dev} & \textbf{Max}
& \textbf{Average} & \textbf{Highest Dev} & \textbf{Max}
& \textbf{Average} & \textbf{Highest Dev} & \textbf{Max} \\
\midrule
\textbf{deepseek-ai/DeepSeek-R1-Distill-Qwen-1.5B}
& 0.224 & 0.332 & 0.342
& 0.190 & 0.299 & 0.299
& 0.572 & 0.661 & 0.669 \\

\textbf{deepseek-ai/DeepSeek-R1-Distill-Qwen-7B}
& 0.483 & 0.528 & 0.537
& 0.385 & 0.429 & 0.439
& 0.634 & 0.660 & 0.691 \\ 

\textbf{google/gemma-2b}
& 0.534 & 0.571 & 0.585
& 0.583 & 0.608 & 0.615
& 0.746 & 0.793 & 0.801 \\

\textbf{google/gemma-2-2b}
& 0.569 & 0.617 & 0.629
& 0.574 & 0.602 & 0.608
& 0.816 & 0.853 & 0.860 \\

\textbf{google/gemma-7b}
& 0.630 & 0.674 & 0.683
& 0.567 & 0.594 & 0.607
& 0.847 & 0.838 & 0.862 \\ 

\textbf{meta-llama/Llama-3.1-8B-Instruct}
& 0.632 & 0.653 & 0.673
& 0.381 & 0.462 & 0.463
& 0.326 & 0.637 & 0.643 \\

\textbf{meta-llama/Llama-3.2-1B-Instruct}
& 0.561 & 0.590 & 0.608
& 0.429 & 0.448 & 0.457
& 0.639 & 0.679 & 0.685 \\

\textbf{meta-llama/Llama-3.2-3B-Instruct}
& 0.570 & 0.616 & 0.625
& 0.434 & 0.460 & 0.462
& 0.779 & 0.795 & 0.801 \\

\textbf{Qwen/Qwen2.5-0.5B}
& 0.263 & 0.365 & 0.368
& 0.384 & 0.415 & 0.426
& 0.527 & 0.551 & 0.557 \\

\textbf{Qwen/Qwen2.5-1.5B}
& 0.517 & 0.537 & 0.546
& 0.446 & 0.457 & 0.467
& 0.695 & 0.749 & 0.755 \\

\textbf{Qwen/Qwen2.5-3B}
& 0.526 & 0.564 & 0.572
& 0.476 & 0.481 & 0.491
& 0.734 & 0.754 & 0.766 \\

\textbf{Qwen/Qwen2.5-7B}
& 0.625 & 0.647 & 0.665
& 0.442 & 0.453 & 0.460
& 0.774 & 0.795 & 0.800 \\
\bottomrule
\end{tabular}%
}
\end{table*}

\begin{table*}[ht!]
\centering
\caption{Generation Tasks: Accuracy metrics for \textbf{gsm8k}, \textbf{mmlu-pro}, and \textbf{math} (Average, Highest Dev, Max).}
\label{tab:generation-tasks}
\resizebox{\textwidth}{!}{%
\begin{tabular}{l
                ccc
                ccc
                ccc}
\toprule
& \multicolumn{3}{c}{\textbf{gsm8k}}
& \multicolumn{3}{c}{\textbf{mmlu-pro}}
& \multicolumn{3}{c}{\textbf{math}} \\
\cmidrule(lr){2-4}
\cmidrule(lr){5-7}
\cmidrule(lr){8-10}
\textbf{Model}
& \textbf{Average} & \textbf{Highest Dev} & \textbf{Max}
& \textbf{Average} & \textbf{Highest Dev} & \textbf{Max}
& \textbf{Average} & \textbf{Highest Dev} & \textbf{Max} \\
\midrule
\textbf{deepseek-ai/DeepSeek-R1-Distill-Qwen-1.5B}
& 0.592 & 0.633 & 0.651
& 0.173 & 0.188 & 0.207
& 0.113 & 0.163 & 0.171 \\

\textbf{deepseek-ai/DeepSeek-R1-Distill-Qwen-7B}
& 0.683 & 0.727 & 0.737
& 0.285 & 0.307 & 0.326
& 0.124 & 0.185 & 0.192 \\

\textbf{google/gemma-2b}
& 0.135 & 0.138 & 0.160
& --    & --    & --
& --    & --    & --    \\

\textbf{google/gemma-2-2b}
& 0.202 & 0.209 & 0.234
& --    & --    & --
& --    & --    & --    \\

\textbf{meta-llama/Llama-3.1-8B-Instruct}
& 0.658 & 0.669 & 0.696
& 0.337 & 0.346 & 0.373
& 0.200 & 0.203 & 0.224 \\

\textbf{meta-llama/Llama-3.2-1B-Instruct}
& 0.291 & 0.298 & 0.325
& 0.140 & 0.140 & 0.168
& 0.095 & 0.109 & 0.123 \\

\textbf{meta-llama/Llama-3.2-3B-Instruct}
& 0.532 & 0.541 & 0.567
& 0.264 & 0.277 & 0.301
& 0.196 & 0.206 & 0.230 \\

\textbf{Qwen/Qwen2.5-0.5B}
& 0.209 & 0.214 & 0.237
& --    & --    & --
& --    & --    & --    \\

\textbf{Qwen/Qwen2.5-1.5B}
& 0.362 & 0.368 & 0.390
& --    & --    & --
& --    & --    & --    \\

\textbf{Qwen/Qwen2.5-3B}
& 0.438 & 0.465 & 0.481
& --    & --    & --
& --    & --    & --    \\

\textbf{Qwen/Qwen2.5-7B}
& 0.589 & 0.613 & 0.625
& --    & --    & --
& --    & --    & --    \\
\bottomrule
\end{tabular}%
}
\end{table*}

\subsection{Development‐Selected Prompt Ordering Algorithm}
Algorithm~\ref{alg:find-order} gives the pseudocode for our development‐selected ordering procedure.

\begin{algorithm}[ht]
  \caption{Development‐Selected Prompt Ordering}
  \label{alg:find-order}
  \begin{algorithmic}[1]
    \Require Demo set \(S_i\), development set \(D_{\mathrm{dev}}\), test set \(D_{\mathrm{test}}\)  
             permutations \(P=128\)
    \For{each demo set \(S_i\) (repeat \(M=10\) times)}
      \State Generate \(\{\pi_j\}_{j=1}^P\) uniformly at random
      \State Evaluate \(a_j = \mathrm{Acc}(S_i,\pi_j\mid D_{\mathrm{dev}})\)
      \State \(\pi^* \gets \arg\max_j a_j\)
      \State Measure \(a^* = \mathrm{Acc}(S_i,\pi^*\mid D_{\mathrm{test}})\)
      \State Measure \(a_{\max} = \max_j \mathrm{Acc}(S_i,\pi_j\mid D_{\mathrm{test}})\)
    \EndFor
    \State \Return \(\{a^*, a_{\max}\}\)
  \end{algorithmic}
\end{algorithm}

\begin{table*}[t]
  \centering
  \small
  \caption{Cross-dataset ordering transfer performance (8-shot generation).}
  \label{tab:cross-transfer}
  \vspace{-2mm}
  \resizebox{\textwidth}{!}{%
  \begin{tabular}{lccccccc}
    \toprule
    \textbf{Model} & \textbf{GSM8K Best} & \textbf{MATH Best} & \textbf{GSM8K Avg} & \textbf{MATH Avg} & \textbf{GSM8K$\rightarrow$MATH} & \textbf{MATH$\rightarrow$GSM8K} & \textbf{Transfer Ratio} \\
    \midrule
    Qwen2.5-0.5B & 0.228 & 0.142 & 0.208 & 0.120 & 0.118 & 0.207 & 0.872 \\
    Qwen2.5-1.5B & 0.378 & 0.148 & 0.360 & 0.119 & 0.109 & 0.355 & 0.837 \\
    Qwen2.5-7B & 0.616 & 0.244 & 0.590 & 0.211 & 0.207 & 0.593 & 0.905 \\
    Qwen2.5-3B & 0.470 & 0.318 & 0.436 & 0.297 & 0.290 & 0.439 & 0.923 \\
    DeepSeek-R1-Distill-Qwen-7B & 0.260 & 0.131 & 0.166 & 0.088 & 0.076 & 0.168 & 0.612 \\
    DeepSeek-R1-Distill-Qwen-1.5B & 0.523 & 0.232 & 0.467 & 0.160 & 0.161 & 0.466 & 0.792 \\
    Llama-3.2-1B-Instruct & 0.266 & 0.029 & 0.232 & 0.009 & 0.012 & 0.234 & 0.644 \\
    Llama-3.1-8B-Instruct & 0.685 & 0.298 & 0.648 & 0.244 & 0.259 & 0.646 & 0.906 \\
    Llama-3.2-3B-Instruct & 0.523 & 0.150 & 0.490 & 0.082 & 0.069 & 0.488 & 0.695 \\
    \midrule
    \textbf{Average} & 0.439 & 0.188 & 0.400 & 0.148 & 0.145 & 0.400 & 0.798 \\
    \bottomrule
  \end{tabular}}
  \vspace{-2mm}
\end{table*}

\end{document}